\documentclass[conference]{IEEEtran}
\IEEEoverridecommandlockouts
\usepackage{cite}
\usepackage{amsmath,amssymb,amsfonts}
\usepackage{algorithmic}
\usepackage{graphicx}
\usepackage{textcomp}
\usepackage{xcolor}
\usepackage{physics}

\def\BibTeX{{\rm B\kern-.05em{\sc i\kern-.025em b}\kern-.08em
    T\kern-.1667em\lower.7ex\hbox{E}\kern-.125emX}}
\begin{document}

\title{Fourier-RNNs for Modelling Noisy Physics Data\\}

\author{\IEEEauthorblockN{Vignesh Gopakumar}
\IEEEauthorblockA{UK Atomic Energy Authority \\
Oxford, United Kingdom \\
\textit{vignesh.gopakumar@ukaea.uk}}
\and
\IEEEauthorblockN{Stanislas Pamela}
\IEEEauthorblockA{UK Atomic Energy Authority \\
Oxford, United Kingdom \\
\textit{stanislas.pamela@ukaea.uk}}
\and
\IEEEauthorblockN{Lorenzo Zanisi}
\IEEEauthorblockA{UK Atomic Energy Authority \\
Oxford, United Kingdom \\
\textit{lorenzo.zanisi@ukaea.uk}}
}

\maketitle

\begin{abstract}
Classical sequential models employed in time-series prediction rely on learning the mappings from the past to the future instances by way of a hidden state. The Hidden states characterise the historical information and encode the required temporal dependencies. However, most existing sequential models operate within finite-dimensional Euclidean spaces which offer limited functionality when employed in modelling physics relevant data. Alternatively recent work with neural operator learning within the Fourier space has shown efficient strategies for parameterising Partial Differential Equations (PDE). In this work, we propose a novel sequential model, built to handle Physics relevant data by way of amalgamating the conventional RNN architecture with that of the Fourier Neural Operators (FNO). The Fourier-RNN allows for learning the mappings from the input to the output as well as to the hidden state within the Fourier space associated with the temporal data. While the Fourier-RNN performs identical to the FNO when handling PDE data, it outperforms the FNO and the conventional RNN when deployed in modelling noisy, non-Markovian data. \\

\end{abstract}


\section{Introduction}
\label{introduction}
There exists an analogy across how neural networks approximates nonlinear functions and how Fourier series represents periodic functions. Both represent parametric methods of estimating the respective functions, neural networks by way of their weights, and Fourier Series by way of the Fourier coefficients. Analogous to the Universal Approximation theorem \cite{Hornik1989}, which defines a neural network's capability in modelling any nonlinearity given an infinitely parameterised network, a Fourier series can model any periodic function given it can be expanded into an infinite sums of sines and cosines. While a neural network parameters are tuned to an optimum value by way of gradient descent, the optimum coefficients for the corresponding periodic function is derived using Fourier Analysis. In practice both the approximation capabilities of neural networks and Fourier series are limited/truncated by the limited parameters that we allocate to it. 

Thus, when employing neural networks to model nonlinearities with heavy periodic dependence, an integrated approach that incorporates nonlinear activation across weights learned in the Fourier space have shown great promise \cite{li2021fourier} \cite{ngom2021fourier}. Even within the scope of NLP,  Fourier transform embedded neural network approaches have become influential \cite{leethorp2021fnet}. Taking the analogy aside these methods have shown considerable performance mainly because they shift the learning regime from the finite dimensional euclidean spaces to that of infinite dimensional spaces using neural operators \cite{cao2021choose}. As demonstrated in \cite{li2021fourier}, Neural Operators allow for mapping from one function space to another function space, allowing it to not be confined by finite dimensional vector mapping. 

Since the introduction of Recurrent Neural Networks (RNNs) \cite{rumelhart:errorpropnonote}, they have been the foundational benchmark for sequential models handling time-series data. Though they have undergone extensive development with more advanced models such as the Gated Recurrent Unit and the Long Short-Term Memory becoming more dominant in applications, the fundamental influence of having separate weight matrices to learn (or unlearn) the hidden state has remained the same across them \cite{cho2014properties} \cite{lstm}. The networks learn not just the forward mapping from the input to the output, but with the aid of the hidden state that is fed back to the network in an auto-regressive manner, it learns the impact the past time instances has on the future time instances, allowing for exploiting the temporal dependencies within the data. 

Through the course of this work, we explore the impact of the amalgamation of these two ideas: integrating neural operator learning within the Fourier space for function mapping with dedicated hidden states fed in recurrently exploiting the temporal dependencies. Though our combined approach,  Fourier-Recurrent Neural Networks (F-RNNs) outperforms standard RNNs in any scenario, they are only comparable to the standard FNO approach when solving for Markovian data as characterised by PDEs. But when they are deployed in modelling noisy, non-Markovian Physics relevant data, they outperform both the classical RNNs as well as the FNO in performance and speed.

\section{Background}
Recent research developments within modelling time-series data has been to employ structures that can exploit the sequential nature of the data. Initially demonstrated through the work on the RNN built by Elman \cite{elman1990finding}, the models have quickly adapted to cover up the flaws within the initial recurrent models. Additional gates added to the architecture of RNNs allowed for longer time extrapolation as demonstrated with the LSTM \cite{lstm}. Currently given adequate data and training resources, with the impact of the self-attention utility, Transformers have become the state-of-the-art model used to model sequential data \cite{Transformer2017}. Temporal nature of the data is often observed with periodic, modal behaviour and analysed using Fourier transform. Taking account the information distribution outlined by the Fourier transform, periodic activation functions have been used within neural networks \cite{SIREN} \cite{Gashler2014}. They have also been used as filtering tools to help neural networks learn higher order features in low-dimensional settings \cite{fourier_features}. Prior research demonstrates that operating within the complex space across the information processing of neural networks demonstrates considerable aid in modelling time-series data \cite{complex_weights}. Extraction of the Fourier basis functions and embedding within the recurrent model is not novel and has been demonstrated to perform better than regular RNNs \cite{FRU}. Our work builds upon all of this previous work and differentiates itself from the Fourier Recurrent Unit (as shown in \cite{FRU}) as it learns the desired behaviour through weighted activations of the inputs in Fourier space rather than via an additive Fourier transform across the input space. 

to be able to extrapolate for longer sequences as shown space has undergone Keeping in mind the sequential nature of time series data, often research developments in modelling efforts have explor been in exploiting the relational context of the inputs.  models that cater to it have built have focussed on exploiting this 
\section{Fourier Neural Operators}
\begin{figure}[ht]
\vskip 0.2in
\begin{center}
\centerline{\includegraphics[width=\columnwidth]{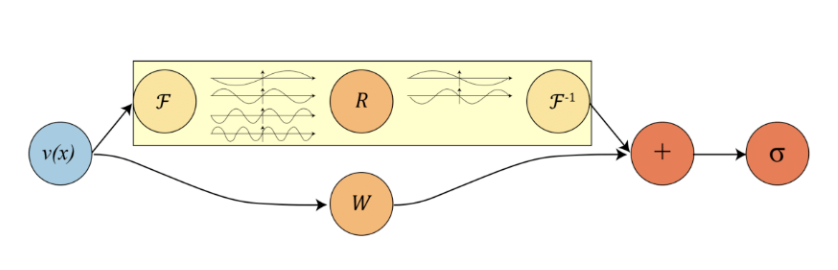}}
\caption{Architecture of the Fourier layer applied within the FNO. It involves two sets of weight matrices $R$ and $W$. $R$ learns the behaviour within the Fourier space while $W$ learns the behaviour within the input Euclidean space. \cite{li2021fourier}}
\label{fourier-layer}
\end{center}
\vskip -0.2in
\end{figure}

The FNO approach shown in \cite{li2021fourier} learns a mapping  between two infinite dimensional spaces from a finite collection of observed input-output pairs. They achieve this by way of deploying a Fourier layer, schematically laid out in Figure \ref{fourier-layer}. 

FNO is constructed by stacking Fourier layers on top of the other. Each Fourier layer is composed of two weight matrices: $R$ and $W$. $R$ learns the mapping behaviour within the Fourier space, while $W$ learns the mapping required within the input Euclidean space. The weight matrix $R$ within the Fourier layer enables convolution within the Fourier space, allowing the network to be parameterised directly in it. The output of a Fourier Layer can be expressed as: 

\begin{equation}
    y = \sigma\bigg(\mathcal{F}^{-1}\big(R\mathcal{F}(x)\big) + Wx\bigg)
    \label{fno_eqn}
\end{equation}

where, 
$x$ is the input, $y$ the output and $\sigma$ is the nonlinear activation function. $\mathcal{F}$ and $\mathcal{F}^{-1}$ represents the Fourier and inverse Fourier transform respectively. 

Throughout the course of this work, we are interested in the \textbf{FNO-2d} configuration as demonstrated in \cite{li2021fourier}, where a 2-d Fourier Neural Operator that only convolves in space is deployed with a recurrent, auto-regressive structure. The FNO takes in as its input the field values across a 2D grid for an initial set of time instances along with the grid discretisation in both dimensions. It outputs a set of later time instances across the desired grid. The outptut of the FNO are then fed back in a recurrent, auto-regressive manner to estimate further field evolution in time.

\section{Recurrent Neural Networks}

RNNs are connectionist models that capture the dynamics of sequences via cycles in the network of nodes \cite{lipton2015critical}. They allow for modelling sequential data, where the output is not just dependent on the input at a given state, but may also depend on the outputs of previous states. They are essentially feed-forward neural networks augmented by the inclusion of edges that span adjacent time steps, introducing a notion of time to the model \cite{lipton2015critical}. This temporal dependency within the model allows RNNs and its different variants effective tools in modeling time dependent physics data. 

\begin{figure}[ht]
\vskip 0.2in
\begin{center}
\centerline{\includegraphics[width=\columnwidth]{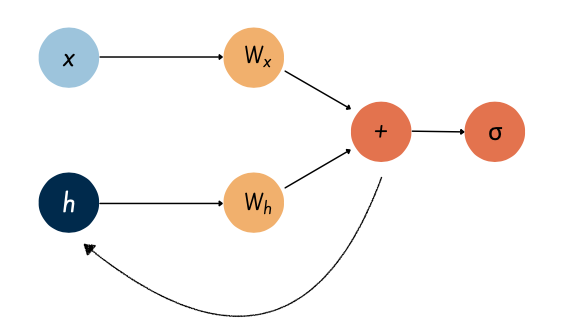}}
\caption{Architecture of an RNN Cell. Each cell is parameterised by two weight matrices: $W_{x}$ and $W_{h}$. The former learns the mappings of the input to the output while the latter learns that impact the hidden state has on the output.}
\label{rnn-cell}
\end{center}
\vskip -0.2in
\end{figure}

Each RNN is composed of RNN Cells stacked on top of each other. The architecture of an RNN cell (as implemented in Pytorch \cite{pytorch}) can be laid out as shown in Figure \ref{rnn-cell}.

Each cell is composed of two linear weight matrices $W_{x}$ and $W_{h}$, tasked with learning different behaviours from the data. $W_{x}$ is tuned to model the function mapping from the input onto the output. It learns the impact that input data at time instance $t$ has on the output at the same instant of time. The linear transformation characterised by $W_{h}$ resembles the hidden state. The hidden state represents the the relevant information from the past time instances that the model has already been exposed to. They can hold sufficient information from an arbitrarily long context window. The functioning of an a RNN Cell can be expressed as: 

\begin{align}
\begin{split}
    h_{t} &= W_{x}x_{t} + W_{h}h_{t-1} \\
    y_{t} &= \sigma(h_t)
\label{rnn_eqn}
\end{split}
\end{align}

where, $x_t$ represents the input at time instant $t$, $h_{t-1}$ the hidden state at time $t-1$, $h_{t}$ the updated hidden state, $y_{t}$ the output at time instant $t$ and $\sigma$ the nonlinear activation function. Biases have been ignored for simplicity. \cite{pytorch}

Each RNN cell takes in two inputs, $x$ and $h$, passes them through linear transformations to update the hidden state of the cell and then through an activation layer to produce the output. The updated hidden state resides within the cell and is fed back in for the next input sequence. Unlike Hidden Markov models, performance of the RNNs are not limited by just looking into the immediate previous state, as hidden states can have deeper memory and look further into the past. 

\section{Fourier-RNNs}
\begin{figure}[ht]
\vskip 0.2in
\begin{center}
\centerline{\includegraphics[width=\columnwidth]{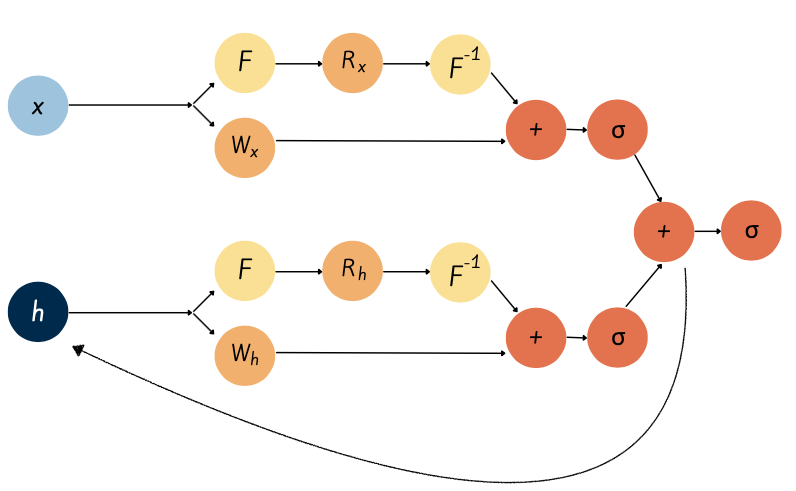}}
\caption{Architecture of an F-RNN Cell. Each cell is parameterised by four weight matrices. $W_{x}$ and $R_{x}$ are trained to map the input to the output in the input space and Fourier space respectively. $W_{h}$ and $R_{h}$ are trained to optimise the hidden state in the input space and Fourier space respectively.}

\label{frnn-cell}
\end{center}
\vskip -0.2in
\end{figure}

Crafting the FNO approach into an RNN we devise the Fourier-RNN, a sequential model capable of performing operator learning for noisy time series data. We construct the F-RNNs by replacing the linear transformation matrices in the RNN with the Fourier layers from the FNO. The architecture of the F-RNN is drawn out in Figure \ref{frnn-cell}. 

The Fourier-RNN is now parameterised by 4 weight matrices as compared to two for the RNN and FNO. Both the input to output mapping as well as the updation of the hidden state happens within the Fourier operator space. These F-RNN cells can be stacked one on top of each other to create F-RNNs with higher expressibility. Mathematically they can be described by replacing equation \ref{fno_eqn} within equation \ref{rnn_eqn}: 

\begin{align*}
    h_t = \mathcal{F}^{-1}\big(R_x\mathcal{F}(x_t)\big) + W_x x_t + \\ \mathcal{F}^{-1}\big(R_h\mathcal{F}(h_{t-1})\big) + W_h h_{t-1} \\
y_{t} = \sigma(h_t)
\label{frnn_eqn}
\end{align*}

Since we have extended the Fourier Neural Operator learning to the hidden state, it is important that we give adequate attention to the intitialisation of the hidden state. Taking into consideration the impact that contextual information has on a recurrent network \cite{wenke2019contextual}, the hidden state at $t_0$ has been modified to accommodate the initial distribution of the field values repeated up to the hidden size and post-fitted with the grid discretisations required by the Fourier operators. 

As opposed to the standard RNN approaches, the F-RNN allows for learning within the Fourier space along with the input space, making it ideal for modelling physics relevant data where modal behavous is seen across space and time. Compared to the FNO, F-RNN is a non-Markovian model, able to account for a longer memory, retaining much more history within the information. FNOs are Markovian and depend on the previous time instance to account for all the required information for predicting the next time iteration. 

\section{Numerical Experiments}
\label{num_expts}

In order to test the efficacy of the F-RNN, we test its performance in modelling the 2D Wave Equation, and the Navier-Stokes Equation at two different Kinematic viscosities, one withi laminar flow and another with turbulent flow. We compare it against the FNO as described in \cite{li2021fourier} and against a Convolutional-RNN (C-RNN). All of the models are trained and tested for noise-free data scenarios as well as for cases with varying degrees of noise. For this purpose, we prepare a synthetic noisy dataset by adding uncorrelated Gaussian noise to the solution data. Noise is sampled from a uniform distribution \((\mathcal{N})\) at mean zero and varying degrees of variance, where we use the variance as the noise factor ($N$). The PDE data corrupted by noise can be expressed as \(\Tilde{x} = x + \mathcal{N}(0, N)\). We initialise a normal distribution and the training and test datasets are augmented by sampling from that distribution. Here, the noise is added to the normalised input and output space. 

For effective comparison, FNO and F-RNN models are built with roughly the same number of parameters (~4.2M) and undergo the same training regime, trained for 1000 epochs with the Adam optimizer with an initial learning rate of 0.001, scheduled to be dropped by a factor of 0.9 every 100 epochs. The C-RNN is built with relatively less (~1.5M) parameters. C-RNNs required a smaller parameter range as larger models were prone to excessive vanishing gradient problem. Exact details about the number of parameters, along with training time for each model for each case can be found in Table \ref{params_table}.  Each dataset comprises of a total 1000 simulations, which is split into 800 for training and 200 for testing. They are fixed onto data loaders with a batch size of 50 epochs. The loss and testing metrics are estimated in mean squared errors and expressed in the unnormalised values. For all of the models, we have chosen the ReLU activation function except for the final activation output of the F-RNN and RNN, for which we have chosen Tanh. For all models a certain number of initial states $(T_{in})$ are fed in to estimate the field at the next time instance (step=1). Each model outputs the next time instance at $T_{in+1}$, which then is fed back into the network along with the previous $T_{in}-1$ instances to estimate the $T_{in}+2$ instance. The cycle continues until we reach the required time instance $T_{out}$. For testing the performance in a noisy scenario, the input data is corrupted with noise and its performance is estimated against the uncorrupted noise-free target data. 

\begin{table}[t]
\caption{Parameter range and average training time (in minutes) for each model. NS$_1$ represents the Navier Stokes equation with $\nu$=1e-3 and NS$_1$ represents the turbulent regime with $\nu$=1e-5}. Each model was trained on a single Nvidia V100 GPU. 
\label{params_table}
\vskip 0.15in
\begin{center}
\begin{small}
\begin{sc}
\begin{tabular}{|c|c|c|c|c|}
\hline
Model & Wave & NS$_1$ & NS$_2$ & Time \\
\hline
C-RNN    & 1079041 & 1614721 & 1603201 & 200  \\
FNO      & 4203873 &  4203873 & 4203553 & 140 \\
F-RNN    & 4201665 &  4201665 & 4201345 & 135 \\

\hline
\end{tabular}
\end{sc}
\end{small}
\end{center}
\vskip -0.1in
\end{table}

For the FNO model, we choose the same architecture as laid out in \cite{li2021fourier}, where the FNO is constructed by stacking 4 Fourier layers. Each layer has a width of 32 for  and is tuned for upto 16 Fourier modes. The C-RNN has an encoded-decoder architecture, where the encoded data, obtained by a series of convolutions is fed into an RNN with 4 hidden layers, each having a size of 256. The output from the RNN is then upsampled to the original dimensions via the decoder using transposed convolutions. The F-RNN consists of two F-RNN cells stacked on top of each other. The hidden size and the width of the Fourier layer is kept at 32, and the model learns upto 16 Fourier modes. 

All experiments were performed on a single Nvidia V100 chip.

\subsection{Navier-Stokes Equation}
We consider the 2D Navier-Stokes equation for a viscous, incompressible fluid in vorticity form: 

\begin{align}
\begin{split}
    \pdv{w}{t} + u.\nabla w &= \nu \nabla^2 w + f \\
    \nabla . u &= 0 
\end{split}
\end{align}

where, $w$ is the vorticity, a function of the velocity field $u$: \(w = \nabla \times u\). They span across a 2D field given by $x$ and $y$, where both lie within the domain $(0,1)$. The time domain for the case is $(0, T_{out})$. $f$ is the forcing function, where $f(x,y) = 0.1((sin(2\pi(x, y)) + cos(2\pi(x, y)))$. We experiment with two kinematic viscosities $\nu$, where $\nu$ = $10^{-3}$ (laminar flow), $10^{-5}$ (turbulent flow). For effective model comparison, the simulation data for the Navier-Stokes case is taken from \cite{li2021fourier}, and the reader is encouraged to peruse the work for more details on the case setup and the numerical solver. 

For the case with $\nu$ = $10^{-3}$, we take in the first 20 time steps (T$_{in}$=20) to the next 20 time steps (T$_{out}$=20). While for that with $\nu$ = $10^{-5}$, we take in the first 10 time steps (T$_{in}$=10) to the next 10 time steps (T$_{out}$=10). For both cases, the step size is 1. 

\begin{figure}[ht]
\vskip 0.2in
\begin{center}
\centerline{\includegraphics[width=\columnwidth]{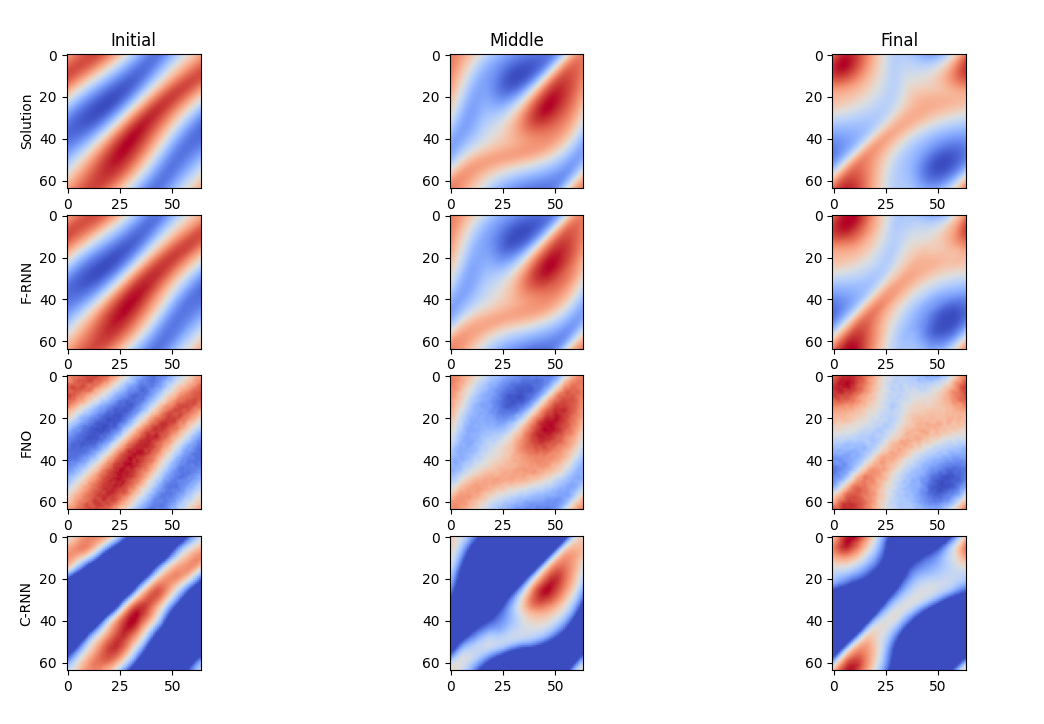}}
\caption{Comparing the performance of each model for the Navier-Stokes case with $\nu$ = $10^{-3}$, where the models take in the first 20 time instances and learn to map the next 20 time instances. Noise factor is set at 0.1. Actual solution is shown at the top, followed by the prediction of the F-RNN and that of the FNO, with the C-RNN at the bottom. The plots shows the network outputs at time t=21 (initial), time=30 (middle), and time=40 (final).}
\label{NS_50_soln}
\end{center}
\vskip -0.2in
\end{figure}

As shown in Figure \ref{NS_50_soln}, it might seem that the performance of the FNO is rather similar to that of the F-RNN, however upon closer inspection we can see that the FNO solution is grainy and corrupted with noise, while the F-RNN is much smoother and closer to the solution. The explicit hidden state, capable of looking longer into the past allows for the F-RNN to understand the impact the noise has across the data and is able to adjust for it within the solution. The FNO, being Markovian and only dependent on the previous state, estimates the noise to be a feature of the solution and is unable to account for it. The impact of the Fourier layers within the F-RNN become clear when compared against the C-RNN methods, which fails to resolve all the modes of operation and is only able to pick up certain dominant behaviours. Experimenting with different noise factors, this is further substantiated with performance variations as shown in Table \ref{ns50-table}. 

\begin{table}[t]
\caption{Performance benchmark at various noise levels for Navier Stokes ($\nu$ = $10^{-3}$). Noise factor represented as $N$.}
\label{ns50-table}
\vskip 0.15in
\begin{center}
\begin{small}
\begin{sc}
\begin{tabular}{|c|c|c|c|c|}
\hline
Model & $N$ = 0.0 & $N$ = 0.05 & $N$ = 0.1 & $N$ = 0.25 \\
\hline
C-RNN    & 0.4789 & 0.4785 & 0.4786 & 0.4791 \\
FNO      & 0.000365 &  0.0006603 & 0.002565 & 0.01368 \\
F-RNN    & 0.0008505 &  0.0009457 & 0.001071 & 0.001499 \\

\hline
\end{tabular}
\end{sc}
\end{small}
\end{center}
\vskip -0.1in
\end{table}
\

Even with the Navier-Stokes case with viscosity $\nu$ = $10^{-5}$, we notice that the F-RNN outperforms FNO and the C-RNN when handling noisy data. Interestingly for this turbulent Navier-Stokes model, the C-RNN fails at all noise levels, and the model gets stuck far from convergence. We believe that within the turbulent regime, the network is unable to differentiate the noise from the fluid behaviour. 

\begin{table}[t]
\caption{Performance benchmark at various noise levels for Navier Stokes ($\nu$ = $10^{-5}$). Noise factor represented as $N$.}
\label{ns20-table}
\vskip 0.15in
\begin{center}
\begin{small}
\begin{sc}
\begin{tabular}{|c|c|c|c|c|}
\hline
Model & $N$ = 0.0 & $N$ = 0.05 & $N$ = 0.1 & $N$ = 0.25 \\
\hline
C-RNN    & 2.137 & 2.137 & 2.137 & 2.137 \\
FNO      & 0.08301 &  0.08792 & 0.09808 & 0.1261 \\
F-RNN    & 0.097 &  0.09234 & 0.09793 & 0.1089 \\
\hline
\end{tabular}
\end{sc}
\end{small}
\end{center}
\vskip -0.1in
\end{table}
\
 The results showcased in Tables \ref{ns50-table} and \ref{ns20-table} shows that our approach, F-RNN performs comparatively with FNO when noise-free or low noise data, but quickly outperforms other models when the noise factor increases.

\subsection{Wave Equation}

Consider the 2D Wave Equation: 

\begin{align}
\begin{split}
    \pdv[2]{u}{t} &= \nu \bigg(\pdv[2]{u}{x} + \pdv[2]{u}{y}\bigg)\\
    \pdv{u}{t} &= 0 \quad ; t = 0
\label{wave_eqn}
\end{split}
\end{align}

where, $u$ is the field value that spans along the $x$ and $y$ axes, both lying within the domain $(-1,1)$. The time domain for the case is $(0, 1)$. The viscosity $\nu$ is taken as 1.0. Periodic boundary conditions are enforced for the problem. The simulation dataset is constructed by varying the initial distribution of the field value. The initial condition for each simulation is a Gaussian given by $u(x,y) = \exp^{-a((x-b)^2 + (y-c)^2)}$, with $a$, $b$ and $c$ are sampled using a Latin Hypercube Sampling \cite{LHS}. The solution for the above equation is built by deploying a spectral solver that uses a leapfrog method for time discretisation and a Chebyshev spectral method on tensor product grid for the spatial discretisation. 

\begin{table}[t]
\caption{Performance benchmark at various noise levels for the 2D Wave Equation. Noise factor represented as $N$.}
\label{wave-table}
\vskip 0.15in
\begin{center}
\begin{small}
\begin{sc}
\begin{tabular}{|c|c|c|c|c|}
\hline
Model & $N$ = 0.0 & $N$ = 0.05 & $N$ = 0.1 & $N$ = 0.25 \\
\hline
C-RNN    & 0.004069 & 0.004656 & 0.005564 & 0.00727 \\
FNO      & 0.001072 &  0.001038 & 0.001116 & 0.001461\\
F-RNN    & 0.0009589 &  0.001064 & 0.001021 & 0.001073\\

\hline
\end{tabular}
\end{sc}
\end{small}
\end{center}
\vskip -0.1in
\end{table}
\

\begin{figure}[ht]
\vskip 0.2in
\begin{center}
\centerline{\includegraphics[width=\columnwidth]{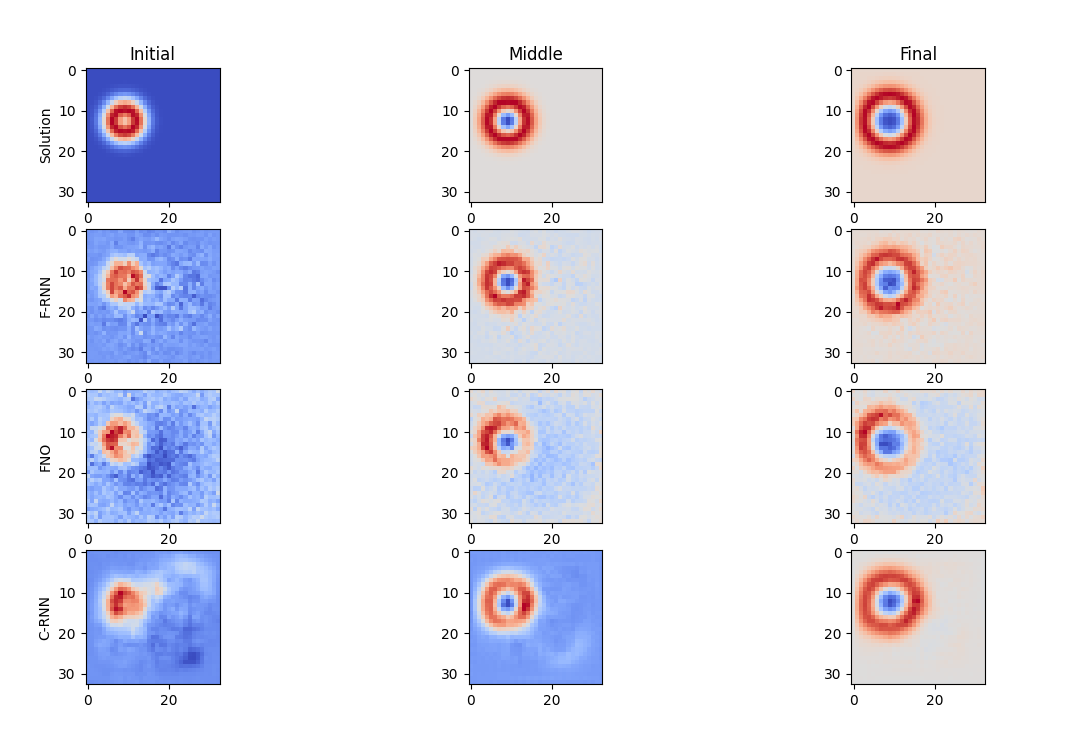}}
\caption{Comparing the performance of each model for the Wave equation, where the models take in the first 20 time instances and learn to map the next 30 time instances. Noise factor is set at 0.25. Actual solution is shown at the top, followed by the prediction of the F-RNN and that of the FNO, with the C-RNN at the bottom. The plots shows the network outputs at time t=0.14 (initial), time=0.24 (middle), and time=0.34 (final). }
\label{wave_soln}
\end{center}
\vskip -0.2in
\end{figure}

The case is setup to take in the first 20 time steps (T$_{in}$=20) to the next 30 time steps (T$_{out}$=30) with a step size of 1, that outputs the next time instance.  

Figure \ref{wave_soln} shows that modelling for 2D wave equation corrupted with a noise factor of 0.25, is a particularly challenging scenario. F-RNN, C-RNN and FNO fail at modelling the exact behaviour. F-RNN being a hybrid between the C-RNN and the FNO, outperforms the latter two models as it extracts the best performing attributes of them both. 

\section{Discussion and Conclusion}

\begin{figure}[ht]
\vskip 0.2in
\begin{center}
\centerline{\includegraphics[width=\columnwidth]{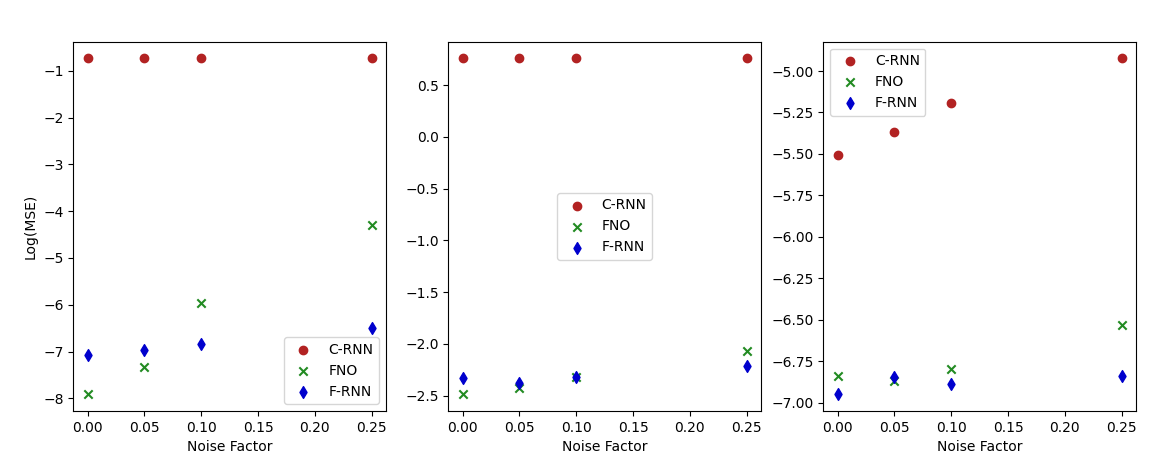}}
\caption{Scatter plot demonstrating the performance of each model at different noise factors. The y-axis represents the logarithm of the mean squared error on the test dataset, while the x-axis indicates the noise factor as mentioned in section \ref{num_expts}. The left figure plots for the Navier-Stokes with laminar flow case, the middle plot represents the Navier-Stokes with turbulent flow case, while the right figure shows the performance for the 2D wave case.}
\label{scatter_plot}
\end{center}
\vskip -0.2in
\end{figure}

Embedding Fourier Layers within an RNN architecture allows for the construction of a more effective sequential model capable of handling noisy time series data, especially physics relevant data where modal behaviour is observed. The Fourier layers are deployed within the F-RNN to map from the input to output space as well as in the updation of the hidden state. This allows for moving from finite-dimensional euclidean mapping spaces to that of infinite dimensions by way of neural operator learning. As can be seen in Figure\ref{scatter_plot}, the performance of the F-RNN is comparable to that of the FNO in low noise scenarios, but as the noise ramps up, F-RNN quickly outperforms the FNO model. When compared against the C-RNN models, F-RNNs are atleast two orders of magnitude more accurate and is much more efficient in GPU time consumption. We believe that our approach has significant value within the surrogate modelling and digital twin space for physical systems and plan to test out the F-RNN on observed experimental data as the next step. 

\newpage
\bibliographystyle{unsrt}
\bibliography{references}

\end{document}